\documentclass[letterpaper, 10 pt, conference]{ieeeconf}  

\IEEEoverridecommandlockouts                          

\usepackage{graphics} 
\usepackage{epsfig} 
\usepackage{mathptmx} 
\usepackage{times} 
\usepackage{amsmath} 
\usepackage{amssymb}  
\usepackage{bm}
\usepackage{dsfont}

\title{\LARGE \bf
EquiDiff: A Conditional Equivariant Diffusion Model For Trajectory Prediction
}

\author{Kehua Chen$^{1,3}$, Xianda Chen$^{2}$, Zihan Yu$^{2}$, Meixin Zhu$^{2,4*}$ and Hai Yang$^{3}$
\thanks{$^{1}$Kehua Chen is with Division of Emerging Interdisciplinary Areas (EMIA), Interdisciplinary Programs Office, The Hong Kong University of Science and Technology, Hong Kong, China.}%
\thanks{$^{2}$Xianda Chen, Zihan Yu and Meixin Zhu are with Intelligent Transportation Thrust, Systems Hub, The Hong Kong University of Science and Technology (Guangzhou), Guangzhou, China.}%
\thanks{$^{3}$Kehua Chen and Hai Yang are with Department of Civil and Environmental Engineering, The Hong Kong University of Science and Technology, Hong Kong, China.}%
\thanks{$^{4}$Meixin Zhu is also with Guangdong Provincial Key Lab of Integrated Communication, Sensing and Computation for Ubiquitous Internet of Things, Guangzhou, China.}
}

\begin{document}

\maketitle
\thispagestyle{empty}
\pagestyle{empty}

\begin{abstract}
Accurate trajectory prediction is crucial for the safe and efficient operation of autonomous vehicles. The growing popularity of deep learning has led to the development of numerous methods for trajectory prediction. While deterministic deep learning models have been widely used, deep generative models have gained popularity as they learn data distributions from training data and account for trajectory uncertainties. In this study, we propose EquiDiff, a deep generative model for predicting future vehicle trajectories. EquiDiff is based on the conditional diffusion model, which generates future trajectories by incorporating historical information and random Gaussian noise. The backbone model of EquiDiff is an SO(2)-equivariant transformer that fully utilizes the geometric properties of location coordinates. In addition, we employ Recurrent Neural Networks and Graph Attention Networks to extract social interactions from historical trajectories. To evaluate the performance of EquiDiff, we conduct extensive experiments on the NGSIM dataset. Our results demonstrate that EquiDiff outperforms other baseline models in short-term prediction, but has slightly higher errors for long-term prediction. Furthermore, we conduct an ablation study to investigate the contribution of each component of EquiDiff to the prediction accuracy. Additionally, we present a visualization of the generation process of our diffusion model, providing insights into the uncertainty of the prediction.
\end{abstract}

\section{INTRODUCTION}
Trajectory prediction is a crucial aspect of autonomous driving, as it plays a significant role in ensuring the safety and efficiency of self-driving vehicles. With the advancement of computer vision techniques, autonomous vehicles are now able to accurately sense their surroundings and make informed predictions about future events. As a result, there has been an increasing interest in developing precise trajectory prediction models for autonomous vehicles. Trajectory prediction models can be classified into several categories, including physics-based, classic machine learning-based, deep learning-based, and reinforcement learning-based methods \cite{huang2022survey}. Among these, deep learning-based models have shown significant promise in achieving accurate predictions. Unlike deterministic deep learning models such as Recurrent Neural Networks (RNNs), attention-based models, and Graph Neural Networks (GNNs), deep generative models aim to learn the data distribution from training data, allowing them to naturally account for uncertainties \cite{gupta2018social}. Therefore, many recent studies have applied deep generative models to predict vehicle trajectories. For instance, Generative Adversarial Networks (GANs) \cite{goodfellow2020generative} and Variational Auto-Encoders (VAEs) \cite{doersch2016tutorial} have been employed to achieve accurate trajectory prediction \cite{ivanovic2020multimodal, gupta2018social}. Moreover, the diffusion model has recently gained popularity due to its powerful capability, and has been applied to trajectory prediction with promising results \cite{gu2022stochastic}.

\begin{figure}[htbp]
\centering
\includegraphics[width=8cm]{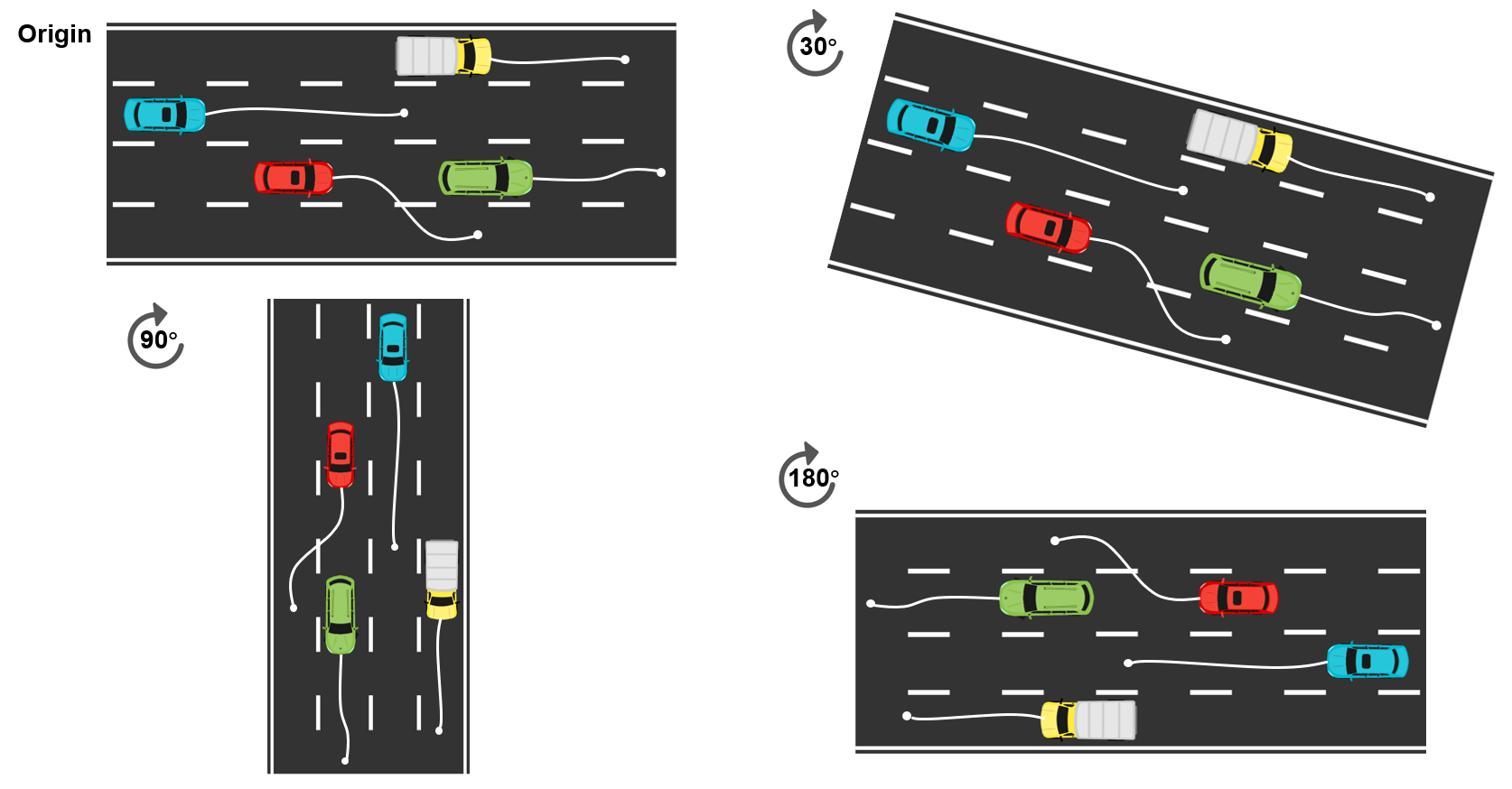}
\caption{Demonstration for Rotation Equivariance.}
\label{fig:rotation}
\end{figure}

Additionally, previous trajectory prediction models do not fully exploit the geometric properties of the trajectory. As 2D position sequences, vehicle trajectories follow geometric equivariant properties. Specifically, we consider SO(2)-invariance and equivariance in this study. As illustrated in Figure \ref{fig:rotation}, when historical trajectories are rotated by different degrees, the relative distances and social interactions among vehicles remain unchanged. However, the relative directions and future trajectories should be rotated by corresponding degrees, i.e., rotation equivariant. Previous studies have typically employed data augmentation techniques to enhance the generalization ability of models with different travel directions. However, data augmentation does not ensure equivariance and requires longer training times. Therefore, recent studies have utilized equivariant properties in trajectory prediction to improve model generalization ability as well as reduce model complexity. For instance, Xu et al. \cite{xu2023eqmotion} proposed a new model, EqMotion, that incorporates equivariant properties into the prediction process.

This paper presents a novel trajectory prediction model, the conditional equivariant diffusion model (EquiDiff). Unlike previous generative models that directly generate future positions of vehicles, we combine the Denoising Diffusion Probabilistic Model (DDPM) with an equivariant transformer to achieve equivariance in the prediction process. Moreover, we design an invariant context encoder to extract historical social interactions among vehicles. To evaluate the performance of EquiDiff, we conduct experiments on the NGSIM dataset, a real-world dataset of highway traffic. 

Our study has the following contributions:
\begin{itemize}
    \item We first combine the diffusion model and equivariant transformer as an equivariant diffusion model for trajectory prediction.
    \item We utilize Graph Neural Networks and Recurrent Neural Networks as a context encoder to extract historical interactions among vehicles.
    \item We verify the performance of our model on a real-world dataset, and show that EquiDiff achieves state-of-the-art accuracy.
\end{itemize}


\section{RELATED WORK}
This section aims to provide an overview of the latest deep learning-based research on predicting vehicle trajectory. Deo et al. \cite{deo2018convolutional} were among the first researchers to utilize LSTM encoder-decoder architecture along with convolutional social pooling to predict future vehicle trajectories, while taking into account social interactions. They began by creating grid cells around the ego vehicle and then generated a social tensor using an LSTM. Finally, a Convolutional Neural Network (CNN) was employed to extract spatial relationships and predict future trajectories. Although CNNs have the ability to capture Euclidean relationships, they are limited in their ability to capture more complex non-Euclidean relationships.
Recently, there has been a growing interest in using graph neural networks (GNNs) to model and predict vehicle trajectories. Li et al. \cite{li2019grip} replaced the convolutional pooling used in Deo et al.'s work with a graph convolutional model to capture more complex spatial relationships. Similarly, Messaoud et al. \cite{messaoud2020attention} applied an attention mechanism pooling to capture historical social interactions and improve trajectory prediction accuracy.

Deep generative models have been increasingly applied in trajectory prediction tasks. Among various deep generative models, adversarial networks have become the most popular framework in recent years. Adversarial networks consist of a generator, which predicts trajectories, and a discriminator, which distinguishes generated trajectories from real ones.
To further improve the accuracy of trajectory prediction, researchers have proposed different variations of adversarial networks. For example, \cite{zhao2019multi} proposed a model called MATF-GAN, which employs multi-agent tensors to capture social interactions among pedestrians. The authors designed an adversarial loss function during training. Another study by \cite{wang2020multi} introduced a novel approach to trajectory prediction using an autoencoder social convolution module and a social recurrent module to fuse spatiotemporal tensors. This model also employs adversarial loss to improve its prediction accuracy. By exploiting the temporal and spatial information of trajectories, the model achieved better performance in predicting trajectories than other models. Additionally, with the thriving of diffusion models, \cite{gu2022stochastic} proposed a novel approach to pedestrian trajectory prediction by employing a diffusion model. The authors utilized a social encoder to generate contextual information, which is then fed to the transformer-based backbone model. The model predicts added noises to generate future pedestrian trajectories.

\section{Preliminary}
\subsection{Denoising Diffusion Probabilistic Model}
Denoising Diffusion Probabilistic Model (DDPM) \cite{ho2020denoising} includes two steps. The first step is called \textit{forward process}, which is a Markov chain. For a sample $\bm{x}_0 \sim q(\bm{x}_0)$, the forward process gradually adds Gaussian noise to the data based on pre-scheduled $\beta_1,...,\beta_K$:
\begin{align}
    q(\bm{x}_{1:K}|\bm{x}_0)=\Pi_{k=1}^Kq(\bm{x}_k|\bm{x}_{k-1})\\
    q(\bm{x}_k|\bm{x}_{k-1})=N(\bm{x}_k;\sqrt{1-\beta_t}x_{k-1},\beta_k\bm{I})
\end{align}

Denote $\alpha_k=1-\beta_k$ and $\bar{\alpha}_k=\Pi_{k=1}^K\alpha_k$, we can derive:
\begin{align}
    q(\bm{x}_k|\bm x_0) = N(\bm x_k; \sqrt{\bar{\alpha}_k}\bm x_0, (1-\bar{\alpha}_k)\bm I)
\end{align}

The second step is \textit{reverse process}, which is also a Markov chain with learned Gaussian starting at $p(\bm x_K)=N(\bm x_K;0, \bold I)$:
\begin{align}
    p_\theta(\bm x_{0:K})=p(\bm x_K)\Pi_{k=1}^Kp_\theta(\bm x_{k-1}|\bm x_k)\\
    p_\theta(\bm x_{k-1}|\bm x_k) = N(x_{k-1};\mu_\theta(\bm x_k, k), \Sigma_\theta(\bm x_k,k))
\end{align}

Afterwards, we can train the model by optimizing the variational lower bound:
\begin{align}
    -\text{log}p_\theta(\bm x_0) &\leq -\text{log}p_\theta(\bm x_0|\bm x_1) + \text{KL}(q(\bm x_K|\bm x_0)||p(\bm x_K)) \nonumber\\
    & + \sum_{k=2}^K \text{KL}(q(\bm x_{k-1}|\bm x_k, \bm x_0)||p_\theta(\bm x_{k-1}|\bm x_k))
\end{align}

In \cite{ho2020denoising}, the authors simplified the objective function as: 
\begin{align}
    L(\theta) = \mathds{E}_{\epsilon, \bm x_0, k}||\epsilon - \epsilon_\theta(\sqrt{\bar{\alpha}_t}\bm x_0 + \sqrt{1-\bar{\alpha}_t}\epsilon, k)||^2
\end{align}
where $\epsilon\sim N(0, \bold I)$.

\subsection{Equivariance and Invariance}
In this study, we aim to develop a model with SO(2) equivariance, i.e. rotation equivariance. Formally, for any rotation matrix $\bm R\in SO(2)$, we have:
\begin{align}
    f(\bm x \bm R) = f(\bm x)\bm R
\end{align}

Similarly, for invariance property, we have:
\begin{align}
    f(\bm x \bm R) = f(\bm x)
\end{align}



\subsection{SO(2)-Equivariant Transformer}
In order to develop a SO(2)-equivariant diffusion model, we employ vector neuron (VN) framework \cite{deng2021vector} and VN-Transformer \cite{assaad2022vn} in this study. Consider a trajectory $\bm x\in \mathds{R}^{T\times 2}$, for linear layers in VN, we use vector neuron representation to process the trajectory as:
\begin{align}
    \bm x' = f_{lin}(\bm x; \bold W)=\bold W\bm x \in \mathds{R}^{C\times 2}
\end{align}
where $\bold W$ is the learnable parameters in the shape of $\bold W\in \mathds{R}^{C\times T}$. 

For non-linear layers with ReLU activation function, we have:
\begin{align}
    &\bold q = \bold W \bm x, \bold k = \bold U \bm x\\
    &x' =
    \left\{
             \begin{array}{lr}
             \bold q,\ \text{if $\langle \bold q, \bold k \rangle \geq 0$}  \\
             \bold q-\langle \bold q, \frac{\bold k}{||\bold k||_2} \rangle \frac{\bold k}{||\bold k||_2},\ \text{otherwise.}  
             \end{array}
\right.\\
& \bm x' = [x']_{c=1}^C = f_{relu}(\bm x)
\end{align}
where $\bold W, \bold U\in \mathds{R}^{1\times C}$ are learnable parameters. With both VN-Linear and VN-Nonlinear, we can have SO(2)-equivariant VN-MLP.

In SO(2)-equivariant transformer, the authors proposed a rotation-equivariant attention mechanism. In detail, they apply Frobenius inner product to compute attention scores. Consider two tensors $\bold Q\in \mathds{R}^{M\times C\times 2}, \bold K\in \mathds{R}^{N\times C\times 2}$, we have:
\begin{align}
    \bold A(\bold Q, \bold K)^{(m)} = \text{softmax}(\frac{1}{\sqrt{2C}}[\langle \bold Q^{(m)}, \bold K^{(n)}\rangle_F]_{n=1}^N)
\end{align}
Then for a value tensor $\bold Z\in \mathds{R}^{N\times C\times 2}$, we have:
\begin{align}
    \text{VN-Attn($\bold Q, \bold K, \bold Z$)}^{(m)} = \sum_{n=1}^N \bold A(\bold Q, \bold K)^{(m, n)}\bold Z^{(n)}\in \mathds{R}^{M\times C\times 2}
\end{align}

\begin{figure}[htbp]
\centering
\includegraphics[width=3cm]{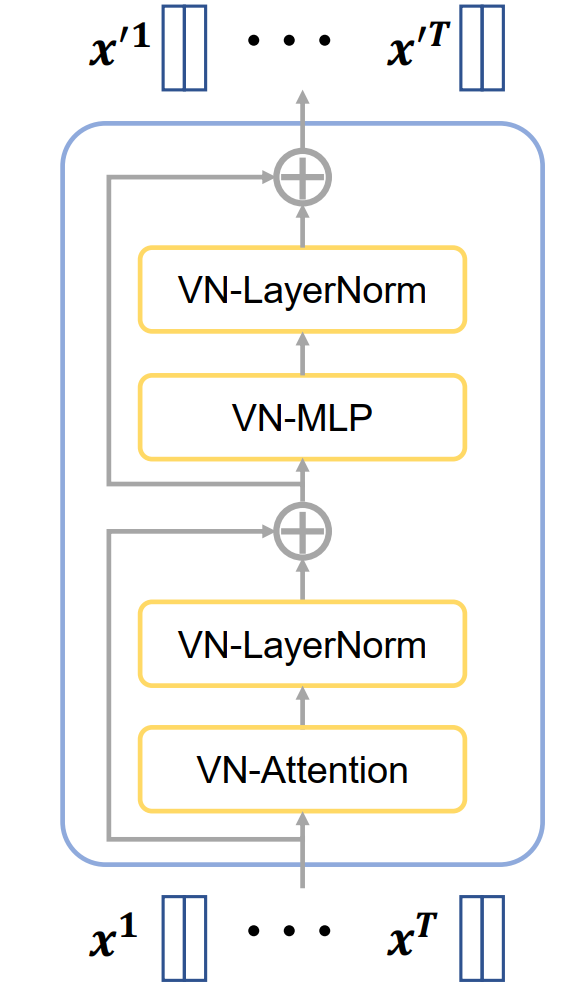}
\caption{Structure of VN-Transformer block.}
\label{fig:vn}
\end{figure}

The layer normalization is then modified as:
\begin{align}
    \text{VN-LayerNorm}(\bold Z^{(n)}) = &[\frac{\bold Z^{(n,c)}}{||\bold Z^{(n,c)}||_2}]_{c=1}^C\odot \nonumber\\
    &\text{LayerNorm}([||\bold Z^{(n,c)}||_2]_{c=1}^C)
\end{align}
where LayerNorm indicates the original layer normalization \cite{ba2016layer}. Figure \ref{fig:vn} presents the structure of VN-Transformer block.

\subsection{Problem Formulation}
The aim of vehicle trajectory prediction is to generate future trajectories for vehicles based on historical trajectories. Formally, given a vehicle $i$ with a historical trajectory $\bm x_i=\{p^t_i\in \mathds{R}^2\}_{t=1}^{T_{his}}$, where $p^t_i$ is 2D coordinates at timestamp $t$, ranging from $t=1$ to $t=T_{his}$. Then we can predict the future movements as $\bm y_i=\{p^t_i\in \mathds{R}^2\}_{t=T_{his}+1}^{T_{pre}}$, the time span ranges from $T_{his}$ to $T_{pre}$. In addition, the trajectory also depends on neighboring vehicles due to complex social interactions. Assume there are $N$ neighboring vehicles around vehicle $i$ denoted as $\mathcal{X}_i\in \{\bm x_j\}_{j=1}^N$, hence, we can utilize the interaction process among vehicles to further improve the prediction accuracy. 

\section{Methodology}
\begin{figure*}[htbp]
\centering
\includegraphics[width=12cm]{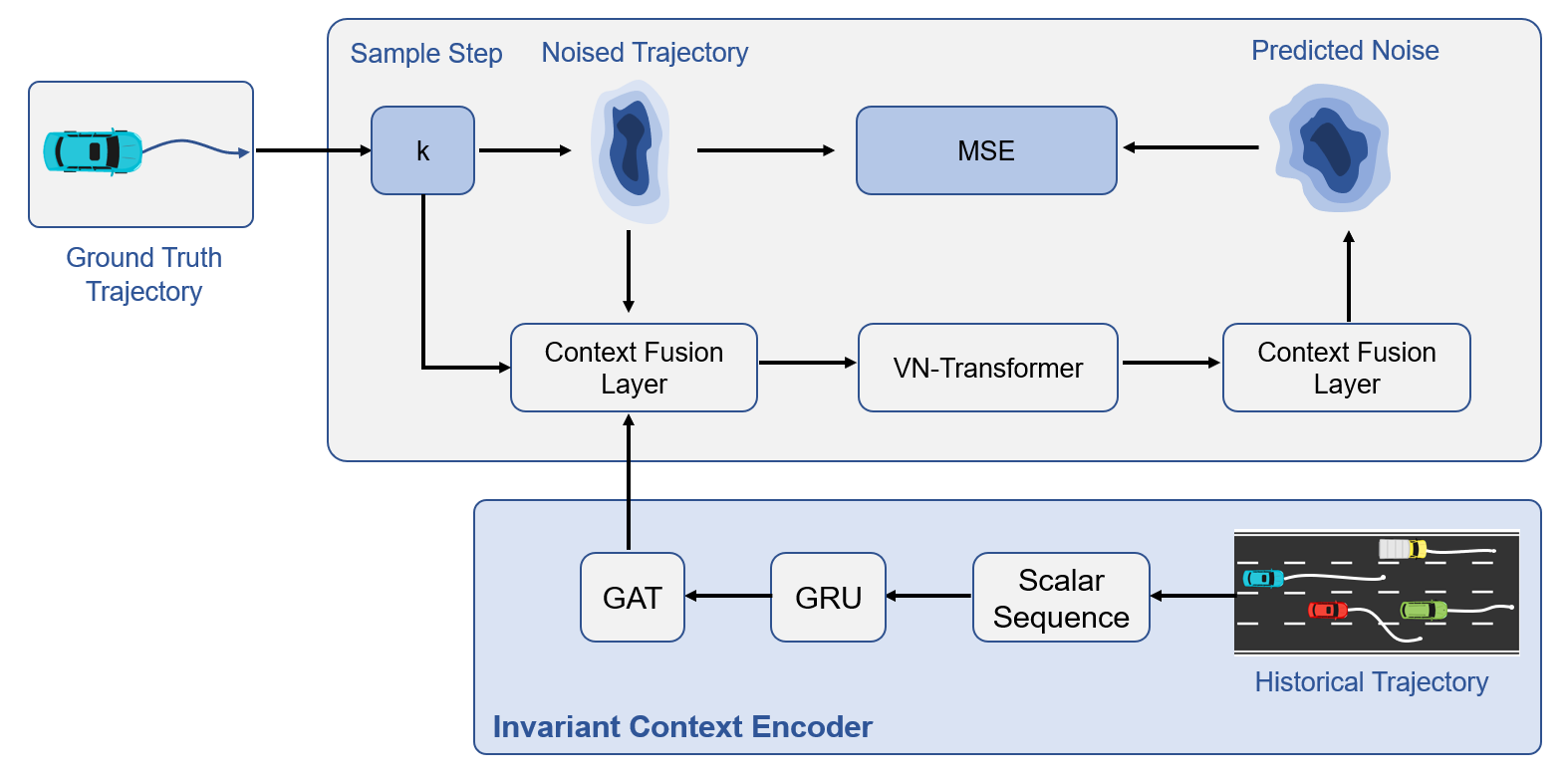}
\caption{Structure of EquiDiff. EquiDiff consists of an invariant context encoder for the extraction of historical social interactions, and a backbone model including context fusion layers and VN-Transformer.}
\label{fig:model}
\end{figure*}

In this section, we introduce the model structure of the proposed EquiDiff (Figure \ref{fig:model}), which is the combination of DDPM and SO(2)-transformer. Different from other auto-regressive models \cite{rasul2021autoregressive}, Equidiff directly generates the whole trajectories. Our model uses VN-Transformer together with context fusion layers as the backbone model to predict the added noise. We further apply an invariant context encoder to extract invariant social interactions among vehicles as the condition.
In detail, we first introduce the invariant context encoder, followed by the backbone model and inference process.

\subsection{Context Encoder}
We implement conditional DDPM in this study. It is easy to understand that future trajectories are conditioned on historical trajectories and social interactions. Hence, the model takes both $\bm x_{k-1}$ and context information $\bm c$ to generate $\bm x_k$ during the sampling process. To extract historical interaction processes among vehicles, we design a context encoder based on Gated Recurrent Unit (GRU) \cite{chung2014empirical} and Graph Attention Network (GAT) \cite{velickovic2017graph}. 

Specifically, each vehicle is deemed as one node on the interaction graph. We first input the historical trajectories of all vehicles into a GRU model. Since the encoder aims to obtain invariant social interactions, we should keep the node features invariant to the rotation operations. Therefore, we first transfer the trajectories into velocity magnitude and angle sequences:
\begin{align}
    &\bm v_t = \bm x_t - \bm x_{t-1}, \ v_t = ||\bm v_t||_2\\
    &\theta_t = \text{arccos}\frac{\bm v_t\cdot \bm v_{t-1}}{||\bm v_t||_2||\bm v_{t-1}||_2}\\
    &\tilde{\bm v}_t = v_t \oplus \theta_t
\end{align}
where $\oplus$ indicates concatenating operation. Then we feed velocity magnitude and angle sequences into GRU model:
\begin{align}
    &r_t = \sigma(\bold W_r\tilde{\bm v}_t+\bold U_r \bold h_{t-1}+\bold b_r)\\
    &z_t = \sigma(\bold W_z\tilde{\bm v}_t+\bold U_z \bold h_{t-1}+\bold b_z)\\
    &\tilde{\bold h}_t=\text{tanh}(\bold W_h\tilde{\bm v}_t+\bold U_h(r_t\odot \bold h_{t-1}))+\bold b_h\\
    &\bold h_t=z_t\odot \bold h_{t-1} + (1-z_t)\odot \tilde{\bold h}_t
\end{align}
where $r_t$ and $z_t$ are the reset gate and update gate respectively; $\bold W\in \mathds{R}^{D\times 2}, \bold U\in \mathds{R}^{D\times D}$, and $\bold b\in \mathds{R}^D$ are learnable parameters; $D$ is the dimension of the hidden states; $\sigma$ is the sigmoid function; and $\odot$ is the Hadamard product.

The last-generated hidden states are treated as node features, and we then apply a GAT model to integrate social interactions. The basic idea of GAT is to utilize attention mechanism to aggregate neighboring information:

\begin{align}
    &\alpha_{ij} = \frac{\text{exp}(\text{LeakyReLU}\big (\bold a(\bold W \bold h_i\oplus\bold W \bold h_j )\big ))}{\sum_{l\in \mathcal{X}_i}\text{exp}(\text{LeakyReLU}\big (\bold a(\bold W \bold h_i\oplus\bold W \bold h_l )\big ))}\\
    &\bold h^{\prime}_i = \sigma(\frac{1}{P}\sum_{p=1}^P\sum_{j\in \mathcal{X}_i}\alpha_{ij}^p\bold W^p\bold h_j)
\end{align}
where $\bold a \in \mathds{R}^{2C}$ and $\bold W\in \mathds{R}^{C\times D}$ are learnable parameters, $C$ is the dimension of the hidden states; $\mathcal{X}_i$ is the neighboring vehicles of vehicle $i$; $P$ is the number of attention heads.

In this way, we acquire the historical driving context for vehicle $i$, denoted as $\bm c_i$. In the following sections, we will use $\bm c_i$ to guide the learning process of the diffusion model. 

\subsection{SO(2)-Equivariant Backbone Model}
In this study, we use VN-Transformer encoders to be the backbone model as introduced previously. Different from the original VN-Transformer, we further apply rotation-equivariant context fusion layers before and after VN-Transformer. In detail, consider the context and trajectory of vehicle $i$ as $\bm c_i$ and $\bm x_i$, we employ Hadamard product to integrate contextual information:
\begin{align}
    &\bm x_i' = (\bold W\bm c_i)\odot \bm x_i\\
    &\bm x_i'' = \text{VN-Transformer}(\bm x_i')\\
    &\bm x_i''' = (\bold W\bm c_i)\odot \bm x_i''
\end{align}
where $\bm c_i\in \mathds{R}^{D\times 1}$; $\bold W\in \mathds{R}^{T\times D}$ is learnable parameters. It is easy to prove that the context fusion layer is SO(2)-equivariant since the context $\bm c_i$ is invariant. We add the context fusion layer before VN-Transformer to introduce contextual information during the attention process. Then we also use a context fusion layer after the VN-Transformer to further enhance the historical information.

\subsection{Inference Process}
After the training process, we can produce future trajectories based on Gaussian noise and contextual information. According to \cite{ho2020denoising}, we sample $\bm y_K\sim N(0, \bold I)$, and generate the prediction step by step:

\begin{align}
    \bm y_{k-1} = \frac{1}{\sqrt{\alpha_k}}(\bm y_k - \frac{\beta_k}{\sqrt{1-\bar{\alpha}_k}}\epsilon_\theta(\bm y_k, k, \bm c)+\sqrt{\beta_k}\bm z)
\end{align}
where $\bm z$ is a random variable in the standard Gaussian distribution; $\epsilon_\theta$ is the trained model which takes previous prediction $\bm y_k$, diffusion step $k$ and contextual information $\bm c$ as inputs.

\section{EXPERIMENT}
\subsection{Dataset}
In this study, we implement experiments on Next Generation Simulation (NGSIM) dataset \cite{nextgenera}. NGSIM consists of two datasets, i.e. I-80 and US-101, with sampling rate as 10 Hz. Each dataset covers 45 mins with mild, moderate and congested traffic conditions. 

Following previous studies \cite{deo2018convolutional}, we use 75\% data in each traffic condition to train the model and the rest for testing. The length of each trajectory is 8 seconds, and we use the first 3 seconds to predict the next 5 seconds. Besides, we downsample the trajectory with factor 2 to make fair comparisons with previous studies. After treatment, the model receives 15-frame historical trajectories to predict the next 25 frames.

\subsection{Metrics}
In this study, we use the root of mean squared error (RMSE) of the predicted trajectories in the next 5-second horizons. RMSE at time $t$ can be calculated as:
\begin{align}
    \text{RMSE} = \sqrt{(\bm x_i^t - \hat{\bm x}_i^t)^2}
\end{align}
where $\bm x_i^t$ and $\hat{\bm x}_i^t$ are ground truth and predicted positions of vehicle $i$ at time $t$, respectively.

\subsection{Model Settings}
We construct interaction graphs based on the circle of 50 meters around the ego vehicles. The dimension of the hidden state is 128, and we employ a four-layer VN-Transformer in the backbone model. For the diffusion model, the initial and terminal $\beta$ are 1e-4 and 5e-2 respectively, and the total diffusion step is 200. During training, the learning rate is 2e-4. We also use the moving average for evaluation with the decay value as 0.999. Additionally, instead of predicting future positions, we found that predicting position offsets has more stable and better performance.

\subsection{Baseline Models}
We compare several state-of-the-art methods in this paper. Since we utilize the generative model to encompass trajectory uncertainties, we mainly consider recent generative-based models for comparison. 

\begin{itemize}
    \item Constant Velocity (CV): CV merely uses a constant Kalman filter to predict future trajectories.
    \item Vanilla LSTM (V-LSTM): The historical trajectories are fed into a vanilla LSTM to predict future positions.
    \item GAIL-GRU \cite{kuefler2017imitating}: GAIL-GRU is based on a generative adversarial imitation learning model to predict trajectories. 
    \item CS-LSTM \cite{deo2018convolutional}: CS-LSTM is an LSTM model with convolutional social pooling layers.
    \item MATF-GAN \cite{zhao2019multi}: MATF-GAN encodes historical trajectories and context as a multi-agent tenor, then extracts interactions based on convolutional fusion. The model also uses adversarial loss during training.
    \item TS-GAN \cite{wang2020multi}: TS-GAN integrates social convolution module, social recurrent module, and generative adversarial framework to model multi-agent spatial-temporal relations.
\end{itemize}

\subsection{Results}

\begin{figure*}[t]
\centering
\includegraphics[width=9cm]{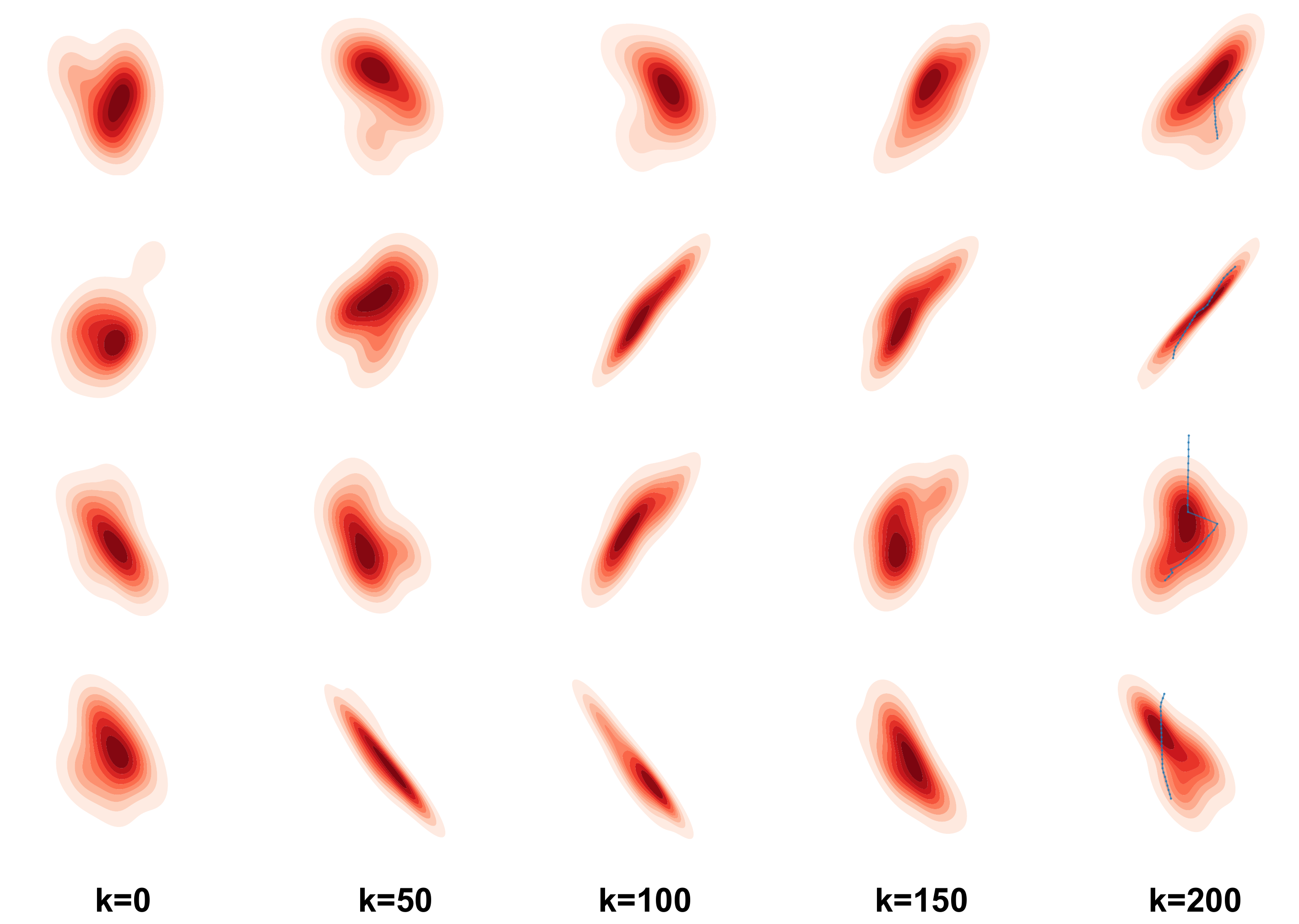}
\caption{Visualization of generation process. The last sub-figures contain the ground truth trajectories via scatters.}
\label{fig:sample}
\end{figure*}

\begin{table}[htpb]
\caption{Comparison of RMSE for 5s Trajectory Prediction}
\centering
\scalebox{1.3}{
\begin{tabular}{c|c|c|c|c|c}
\hline
\textbf{Model}&\textbf{1s}&\textbf{2s}&\textbf{3s}&\textbf{4s}&\textbf{5s}\\
\hline\hline
CV  &0.73&1.78&3.13&4.78&6.68 \\
V-LSTM& 0.68&1.65&2.91&4.46&6.27\\
GAIL-GRU&0.69&1.51&2.55&3.65&4.71\\
CS-LSTM&0.61&1.27&2.09&3.10&4.37\\
MATF-GAN&0.66&1.34&2.08&2.97&4.13\\
TS-GAN&0.60&1.24&1.95&\textbf{2.78}&\textbf{3.72}\\
\hline
EquiDiff-Eq&0.58&1.26&1.95&3.03&4.03\\
EquiDiff-Ctx&0.65&1.41&2.23&3.54&4.55\\
EquiDiff&\textbf{0.55}&\textbf{1.21}&\textbf{1.92}&3.03&4.01\\
\hline
\end{tabular}}
\label{Tab:res}
\end{table}

Table \ref{Tab:res} presents a comparison of the RMSE for various methods. Deep learning-based models have shown to outperform traditional models due to the powerful expressiveness of deep neural networks. While V-LSTM and GAIL-GRU take temporal features into account, they do not utilize interactions among vehicles, and therefore only perform slightly better than the CV method. In contrast, CS-LSTM, MATF-GAN, and TS-GAN all leverage CNNs to extract social interactions. However, MATF-GAN and TS-GAN employ additional adversarial losses during the training process to improve the accuracy of the prediction models. As a result, these models achieve better prediction accuracy compared to CS-LSTM.

In contrast to adversarial models, our EquiDiff approach utilizes a diffusion process to gradually generate future trajectories from an isotropic Gaussian distribution. Additionally, we fully leverage the geometric properties of position coordinates by employing an SO(2)-equivariant transformer. Our experimental results demonstrate that EquiDiff can more accurately predict short-term future trajectories than other generative models. However, it is observed that EquiDiff suffers from higher errors for long-term predictions, specifically for time horizons of 4-5 seconds. This is mainly attributed to the fact that we do not consider multi-modality in this paper, and different maneuvers of drivers can have a significant impact on long-term trajectories. We intend to address this limitation in future work.

\subsection{Ablation Study}
To verify the effectiveness of each component, we conducted an ablation study in this section. Specifically, we introduced two variants for comparison. The first variant eliminates the equivariant transformer, and a vanilla transformer encoder is used, denoted as EquiDiff-Eq. As shown in Table \ref{Tab:res}, EquiDiff-Eq exhibits competitive performance compared with other baseline models, but it still performs worse than EquiDiff. These results suggest that the incorporation of equivariance is essential to capture the geometric features of the trajectory data.
In the second variant, we solely use a GRU to extract the temporal features of the ego vehicle and do not consider social interactions, i.e., the GAT model is eliminated from the context encoder, denoted as EquiDiff-Ctx. As shown in our experimental results, prediction errors significantly increase when social interactions are not considered. This finding is consistent with previous observations and highlights the critical role of vehicle interactions in trajectory prediction tasks.

\subsection{Sampling Visualization}
In this section, we randomly selected four cases for visualization to provide an intuitive understanding of the trajectory prediction generated by our EquiDiff model. Figure \ref{fig:sample} illustrates the trajectory generation process via sampling from $k=0$ to $k=200$. The red distributions indicate the predicted trajectories with uncertainties. Additionally, we show the ground truth trajectories in the last sub-figures (i.e., when $k=200$) for comparison. It can be observed that at the beginning of the sampling process, the joint distribution of position coordinates follows Gaussian distributions. As the sampling steps progress, the generated distributions concentrate on the ground truth trajectories. However, some differences between the prediction distribution and the ground truth trajectories can still be observed, particularly when drivers perform lane-changing or overtaking maneuvers.

\section{CONCLUSIONS}
In this paper, we propose a novel approach for vehicle trajectory prediction by combining the Denoising Diffusion Probabilistic Model (DDPM) with the SO(2)-equivariant transformer. We evaluate the performance of our proposed model on the NGSIM dataset, and the results demonstrate that our model outperforms other baseline models in terms of short-term prediction accuracy. Although our model performs slightly inferior to other generative models for long-term prediction, the combination of the diffusion model and equivariant model is promising and has the potential to achieve state-of-the-art accuracy.




\section*{ACKNOWLEDGMENT}
We would like to acknowledge a grant from RGC Theme-based Research Scheme (TRS) T41-603/20R, and a research grant (project N\_HKUST627/18) from the Hong Kong Research Grants Council under the NSFC/RGC Joint Research Scheme. This study is also funded by the Guangzhou Municiple Science and Technology Project (2023A03J0011); and Guangzhou Basic and Applied Basic Research Project (SL2022A03J00083).

\bibliographystyle{ieeetr}

\bibliography{refs}

\end{document}